# A Machine Learning Model for Crowd Density Classification in Hajj Video Frames

Afnan A.Shah

Department of Computing-College of Engineering and Computing, Umm Al-Qura University, Makkah 21961, Saudi Arabia

*Abstract*—Managing the massive annual gatherings of Hajj and Umrah presents significant challenges, particularly as the Saudi government aims to increase the number of pilgrims. Currently, around two million pilgrims attend Hajj and 26 million attend Umrah making crowd control especially in critical areas like the Grand Mosque during Tawaf, a major concern. Additional risks arise in managing dense crowds at key sites such as Arafat where the potential for stampedes, fires and pandemics poses serious threats to public safety. This research proposes a machine learning model to classify crowd density into three levels: moderate crowd, overcrowded and very dense crowd in video frames recorded during Hajj, with a flashing red light to alert organizers in real-time when a very dense crowd is detected. While current research efforts in processing Hajj surveillance videos focus solely on using CNN to detect abnormal behaviors, this research focuses more on high-risk crowds that can lead to disasters. Hazardous crowd conditions require a robust method, as incorrect classification could trigger unnecessary alerts and government intervention, while failure to classify could result in disaster. The proposed model integrates Local Binary Pattern (LBP) texture analysis, which enhances feature extraction for differentiating crowd density levels, along with edge density and area-based features. The model was tested on the KAU-Smart-Crowd 'HAJJv2' dataset which contains 18 videos from various key locations during Hajj including 'Massaa', 'Jamarat', 'Arafat' and 'Tawaf'. The model achieved an accuracy rate of 87% with a 2.14% error percentage (misclassification rate), demonstrating its ability to detect and classify various crowd conditions effectively. That contributes to enhanced crowd management and safety during large-scale events like Hajj.

## I. Introduction

Mass surveillance video data recorded during Hajj require advanced processing techniques [1] to classify and detect crowd density levels in real-time ensuring timely responses to high-risk situations [2]. Manual crowd-monitoring systems are not only time-consuming and resource-intensive but also insufficient for managing the complexity of massive events like Hajj [3]. Machine learning methods have been implemented as a reliable solution for automating crowd analysis enabling the classification of crowd density levels directly from video frames [4]. Managing crowd density during Hajj also presents general challenges [5], such as noisy video frames [6], environmental factors like poor visibility and unrecognized objects [7], such as moving fans or vehicles near the Kaaba. These challenges can cause false alerts or missed detections [8], potentially leading to delays or dangerous situations. Furthermore, normal crowd behaviors during Hajj, such as group movements [9] and sacred congregations may appear high-risk in other contexts.

This study focuses on detecting and classifying crowd density into three levels: moderate crowd, overcrowded and very dense crowd. Unlike many current research efforts that predominantly rely on Convolutional Neural Networks (CNNs) for detecting abnormal behaviors [10], this research employs a Gradient Boosting Classifier (GBC) with a structured feature extraction model. Features such as Edge Density, Local Binary Pattern (LBP) texture and crowd area coverage provide a comprehensive representation of crowd characteristics, while a threshold-based method directly labels frames with Edge Density values above a critical level as very dense, ensuring the rapid identification of high-risk crowd regions.

The proposed model implements data augmentation techniques to improve generalization throughout various scenarios, uses class balancing to address the distribution of crowd density levels in the dataset and enhances the GBC through hyperparameter tuning with GridSearchCV to ensure robust performance. In addition, the model triggers visual alerts for very dense crowd scenes by overlaying a red flash indicator on detected frames, assisting organizers in real-time decision-making.

The contributions of this study are as follows: development of a Machine Learning model for classifying crowd density levels in video frames recorded during Hajj; integration of key features such as Edge Density, LBP texture and crowd area coverage to enhance classification accuracy; and use of performance metrics such as accuracy, precision, recall and F1-score to evaluate the model, with special emphasis on the 'very dense crowd' category.

The following sections are organized as, Section II provides the related works and Section III presents the proposed model. Section IV shows the result analysis and Section V concludes the paper.

## II. Related Work

Felemban et al. [2] highlighted different levels of crowd density observed among pilgrims categorized as moderate crowd, overcrowded and very dense crowd. Such categories range from safe to potentially disastrous with very dense crowds posing significant risks such as stampedes. Although extensive efforts by Hajj authorities to manage crowds effectively, a method to classify hazardous crowd density conditions and provide timely notifications remains critical for enhancing crowd management. This paper builds on such insights by proposing a data-driven machine-learning approach for accurate and timely crowd density classification.

Aldayri and Albattah [11] introduced a computer vision framework that uses a convolutional LSTM autoencoder to detect abnormal human behaviors in video sequences by combining convolutional neural network (CNN) to extract spatial features from video frames and LSTMs to analyze temporal patterns across consecutive frames. Their research, along with other similar studies [12] focuses on identifying abnormal behaviors whereas my model emphasizes crowd density classification based on structural features like Edge Density and LBP texture, addressing the specific challenge of real-time detection of very dense crowd levels.

In another research effort [13], a Crowd Anomaly Detection Framework (CADF) was developed which integrates multi-scale feature fusion and soft non-maximum suppression (soft-NMS) to detect anomalies in dense crowd scenarios. Bhuiyan [4] proposed the Crowd Anomaly Hajj Monitor to classify crowd anomalies by using optical flow and a fully convolutional neural network, which identifies specific behaviors but counts on predefined feature extraction techniques that may limit adaptability to varying crowd scenarios. On the other hand, my approach leverages Edge Density and other texture-based features for crowd density classification to ensure wider relevance in different crowd conditions. In addition, Alhothali et al. [14] introduced a deep CNN-based model to detect and localize anomalous events in dense crowd scenes classifying seven categories of abnormal events based on extracted spatial and temporal features. While that approach is effective for anomaly detection, such models often focus on individual behaviors rather than holistic crowd density levels. In a similar effort [15], other studies have utilized lightweight CNN and LSTM models to detect violent activities in surveillance footage, generating real-time alarms for law enforcement, which often enhance specific scenarios. Whereas, my study emphasizes a generalizable model for crowd density classification that focuses on critical safety categories like very dense crowds.

Previous works have widely applied CNNs and FCNNs to analyze surveillance videos for detecting anomalies [16]. Also, FCNN have been used for estimating crowd density from distant surveillance footage, showing significant improvements in crowd density classification [17]. However, many depend on fixed feature extraction techniques, such as optical flow or motion analysis [18] which can limit their adaptability to the unique dynamics of Hajj. Also, methods dependent on inflexible frameworks [19] may face challenges in handling the complicated temporal and spatial variations in crowd movement, especially during rituals and gatherings. The proposed model in this paper addresses these limitations by extracting meaningful features such as Edge Density, LBP texture and crowd area coverage which better capture the visual and structural characteristics of different crowd density levels. Furthermore, traditional approaches may classify normal crowd behaviors during Hajj, such as ritual-related movements or group activities as anomalies because of the unique nature of such event. This study avoids such pitfalls and provides a focused solution for monitoring critical crowd density levels in real-time by tailoring the model to classify crowd density rather than individual behaviors.

This study adds to the existing body of knowledge by presenting a machine-learning model designed to systematically classify crowd density levels. The model achieves enhanced accuracy and adaptability by employing a structured feature extraction process and fine-tuning classification with GBC, offering a practical solution for improving crowd management and safety during Hajj.

## III. THE PROPOSED MODEL

### A. Data Preparation

I used the HAJJv2 dataset [20] which contains training and testing videos each set including nine videos capturing different holy sites during Hajj such as Massaa, Jamarat, Arafat and Tawaf. This dataset was originally created to focus on abnormal crowd behaviors during Hajj. Each video is divided into different frames and each frame's number is repeated multiple times in the label file with each instance assigned to a different type of abnormal behavior, thereby allowing for the capture of various behaviors occurring simultaneously or consecutively within the same frame.

For this study, I used the same segmented frames provided in the dataset with each video divided into a varying number of frames ranging between 466 and 501. However, I modified the label files to categorize the frames into three classifications; moderate crowd, overcrowded and very dense crowd eliminating the need to repeat the same frame's number to define multiple abnormal behaviors in a single frame. This was essential to reduce complexity and focus on overall density rather than specific behaviors.

Since the goal of this study is to categorize the overall crowd density in each frame rather than track specific abnormal behaviors or objects, I eliminated the need to repeat the same frame number to detect various abnormal behaviors in different areas within each frame. Instead, I simplified this by selecting one entry per frame to represent its density level. Each video's frames in the training and testing set are processed individually and assigned a single density label; moderate, crowded or very dense crowd based on the dominant or average density level observed in each frame.

To ensure accurate density classification, the dataset was preprocessed further. During this process, the total number of rows loaded from the dataset was 43,721 and frames with high EdgeDensity values were automatically updated with corresponding labels for very dense crowd conditions. The augmented dataset resulted in 131,163 entries which included features such as Edge Density, Area, Average Intensity and Local Binary Pattern (LBP) texture, significantly expanding the variety of training samples. In addition, an edge density threshold of 64.5 was set during the training process to facilitate the reliable identification of very dense crowds. That simplifies the dataset structure, making it more efficient for training a model aimed at detecting general crowd density levels without requiring additional details, such as bounding boxes or multiple labels per frame. By doing so, the training process is more streamlined, with improvements in efficiency and performance.

*B. Model Design*

The major goal of the proposed model is to detect and classify crowd density levels in video frames recorded during Hajj across various holy sites aiming to identify very dense crowd regions that may require immediate attention. In order to achieve this, the model uses a structured approach that starts with careful dataset preparation and well-defined labeling criteria as shown in Fig 1. The dataset comprises video frames categorized by crowd density levels: moderate crowd, overcrowded and very dense crowd. The model employs a threshold-based method to automatically label samples where frames with an EdgeDensity value above 64.5 are directly classified as very dense. This was selected through exploratory analysis of sample distributions and is intended to provide immediate labeling for the most critical category, ensuring timely detection of high-density scenes.

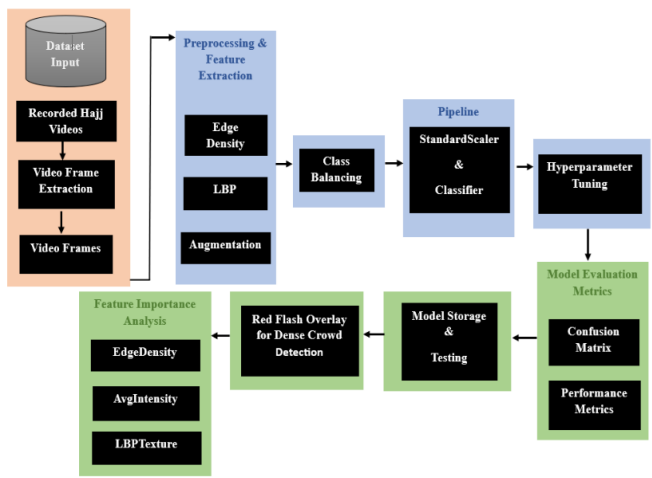

Fig. 1. Proposed model.

The model extracts Edge Density by calculating the density of edges detected in a designated area of each frame as this feature reflects changes in density. The Local Binary Pattern (LBP) texture feature was calculated by analyzing the local pixel patterns in grayscale crowd regions within the video frames, by which the LBP algorithm computes a binary code for each pixel by comparing it with its surrounding neighbors and generates a histogram capturing the frequency of specific binary patterns. That histogram reflects the texture characteristics of the crowd which is particularly useful in identifying varying density levels based on textural patterns that are prevalent in dense crowds. In this study, the histogram values were normalized and averaged to provide a single numerical representation of texture per region, which was included as the LBP Texture feature. The decision to use LBP stems from its ability to highlight subtle texture variations across crowd density levels aiding the model in making more accurate classifications and features like the area covered by the crowd region and average intensity contribute further contextual details. Also, data augmentation is used to enhance the training set's diversity including frame rotation and brightness adjustment. The reason for using augmentation is to provide the model with a wider array of samples thereby improving its generalization across new frames by allowing it to learn from a broader variety of crowd presentations.

Class balancing was essential due to some crowd levels having significantly fewer samples than others. The model addresses this by oversampling the minority classes, particularly the moderate crowd and very dense crowd categories. This ensures that each crowd level is equally represented during training thereby reducing the likelihood of bias and enhancing the classifier's performance across all classes.

The Gradient Boosting Classifier (GBC) was implemented as the primary classification algorithm because of its effectiveness in modeling complex non-linear relationships. That classifier works by sequentially building a series of decision trees where each tree attempts to correct the errors of its predecessor. That iterative boosting process allows the model to focus more on the misclassified samples from previous iterations, thereby improving its overall accuracy. For this study, GBC was integrated into a model that included feature scaling using StandardScaler to normalize the values of all features e.g., EdgeDensity, LBPTexture, Area, and AvgIntensity. The scaled features were then passed to the classifier. The model achieved optimal performance by incorporating GBC in a model and tuning its hyperparameters using GridSearchCV, as reflected in a cross-validation score of 0.97. That iterative approach also ensures the model's robustness in classifying different crowd densities despite the diversity in frame characteristics.

Hyperparameter tuning through GridSearchCV allows for the optimization of the GBC. A selection of key parameters such as the number of estimators, learning rate and maximum depth is tuned using cross-validation to ensure the model performs well without overfitting. Systematically, GridSearchCV tests various parameter options to find the set that yields the highest accuracy score during cross-validation.

Performance metrics such as accuracy, precision, recall and F1-score are used to evaluate the proposed model in this study as they provide an in-depth assessment of its ability to predict each crowd density level correctly with particular attention given to its precision and recall on the 'very dense crowd' classification due to the high importance of this category. In addition, a confusion matrix is employed to analyze further misclassifications offering insights into which crowd density levels may be confused by the model.

The final model, saved as a serialized file, is tested on new video frames to assess its practical performance in detecting very dense crowds, and importantly, it activates a red flash indicator as a visual alert for such frames. The design process is finalized with a feature importance analysis as each feature is examined to provide findings on which factors of crowd data affect the model's predictions validating the choice of features and pointing out the model's reliance on important crowd-related characteristics.

*C. Training Process*

The training dataset was balanced using oversampling techniques to address the class imbalance especially for the moderate crowd and very dense crowd categories. That ensured equal representation of all three crowd density levels. Features of each frame were extracted that included Edge Density calculated as the density of detected edges, Local Binary Pattern (LBP) texture, area covered by the crowd and average intensity.

During the training process, the LBP texture features played a significant role in distinguishing between moderately crowded and very dense crowd frames, since the histogram values captured by LBP provided insights into the structural patterns within the crowd, such as uniform regions indicating low density versus textured regions in high-density areas. That feature combined with EdgeDensity, area, and average intensity, allowed the model to detect high-density areas effectively. In addition, to enhance the diversity of the training set, data augmentation techniques such as rotation and brightness adjustments were employed.

*D. Hyperparameter Optimization and Feature Importance*

Hyperparameter tuning was performed using GridSearchCV, optimizing key parameters of GBC, such as the number of estimators 50, 100 and 200, learning rate 0.01, 0.05, 0.1, 0.3 and 0.5 and maximum depth 3, 4, 5, 6 and 7. The best parameters were a 0.5 learning rate, a maximum depth of 7, and 200 estimators, achieving a cross-validation score of 0.97 which provided the best cross-validation accuracy while balancing model complexity and performance.

An analysis of feature importance revealed that Edge Density was the most significant predictor for crowd density classification followed by LBP Texture and Average Intensity. The feature importance were: Area (0.69), Edge Density (0.24), Average Intensity (0.06) and LBP Texture (0.00). That confirms the relevance of these features in distinguishing between different crowd densities and supports their inclusion in the model.

## IV. RESULTS AND DISCUSSION

During the training stage of the model, I visualized the distribution of key features, such as area, position coordinates, edge density and average intensity across different crowd density levels including moderate, overcrowded and very dense crowd. These highlight how these features contribute to distinguishing between the crowd density categories which is essential for the model's accuracy. Fig. 2 shows that Area is a significant feature in distinguishing very dense crowd situations from moderate crowd or overcrowded as very dense crowds tend to cluster within a smaller area. In the HAJJv2 dataset, very dense crowds mostly appear in videos recorded during Tawaf around the Kaaba which is typically a quite small area. The yellow and green bars representing overcrowded and moderate crowds respectively, are distributed more widely across larger areas such as Sa'i, Mina, and Arafat compared to the very dense category. Such finding suggests that as crowd density decreases, the area covered by the crowd tends to increase due to people being more spread out.

Fig. 3 and Fig. 4 show the distribution of X and Y coordinates respectively, by crowd density category: moderate crowd, overcrowded and very dense crowd. Very dense crowd regions represented by the red bars are concentrated within specific ranges on both the X and Y axes. The clustering suggests that very dense crowds are likely to occur in particular locations within the frame reflecting areas where crowding frequently happens. Meanwhile, overcrowded and moderate crowds represented by the yellow bars and green bars respectively, are more widely distributed across the X and Y coordinates. In particular moderate crowds are spread across a larger range, indicating a more even distribution across the frame. Therefore, the specific positioning patterns of each crowd density level imply that spatial coordinates X and Y are useful features for distinguishing crowd densities. Thus, the model can use this information to recognize crowd density levels based on where they commonly appear within the frame.

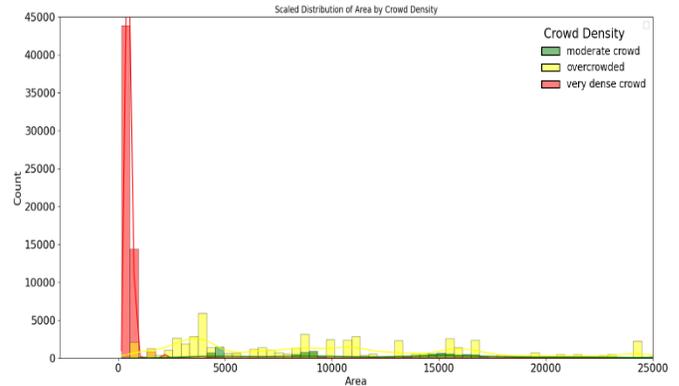

Fig. 2. Distribution of the "Area" feature by crowd density category (moderate crowd, overcrowded, and very dense crowd).

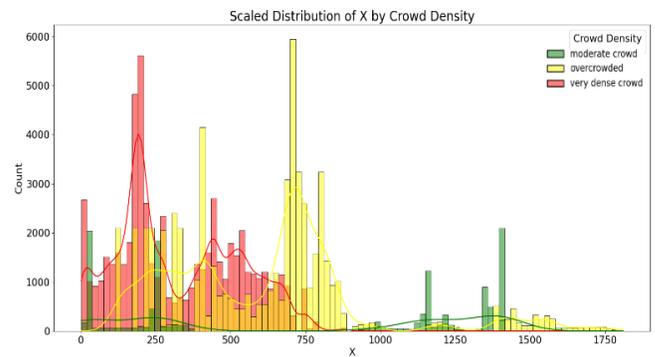

Fig. 3. The distribution of the " X " coordinate by crowd density category.

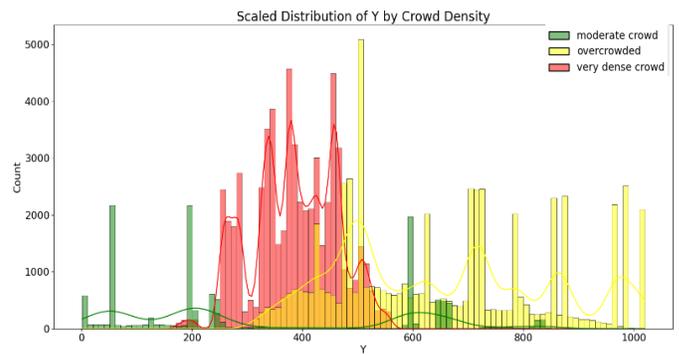

Fig. 4. The distribution of the "Y" coordinate by crowd density category.

For the distribution of the 'Edge Density' feature across the three crowd density categories, Fig. 5 shows that the very dense crowd represented by the red color, has the highest concentration of samples at larger edge density values, with a peak around 80–100.

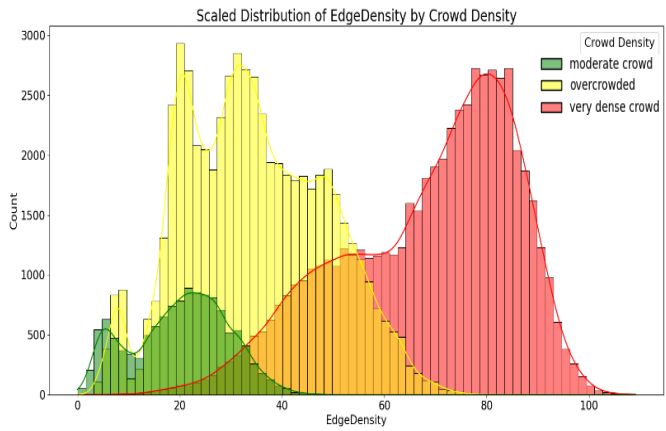

Fig. 5. Distribution of the "Edge Density" feature across crowd density categories.

Additionally, the figure shows that the overcrowded and moderate crowd categories, represented by yellow and green respectively, have a peak in the middle range of edge density values (around 40–60) and are concentrated in the lower range of edge density values below 20, respectively. This highlights the significance of edge density as a feature for classifying crowd density levels, as the distributions show clear separations, especially between the moderate and very dense crowd categories, with minimal overlap. Thus, the 'Edge Density' feature is a major factor contributing to the model's accuracy in distinguishing between these categories.

During the counting of the 'Average Intensity' feature, the visualized values show that, while there are slight shifts in the distributions as shown in Fig. 6, the very dense crowd peaks at slightly higher average intensity values around 125–150, the overcrowded category peaks in the mid-range of 100–125, and the moderate crowd is concentrated at lower intensity values below 100. All three crowd density levels overlap significantly in the range of 75 to 125 average intensity values. In addition, Fig. 5 demonstrates that average intensity captures variations in brightness across the frame that correlate with crowd density. While the overlap limits its standalone utility, it can enhance the model's accuracy when combined with other features like edge density and area.

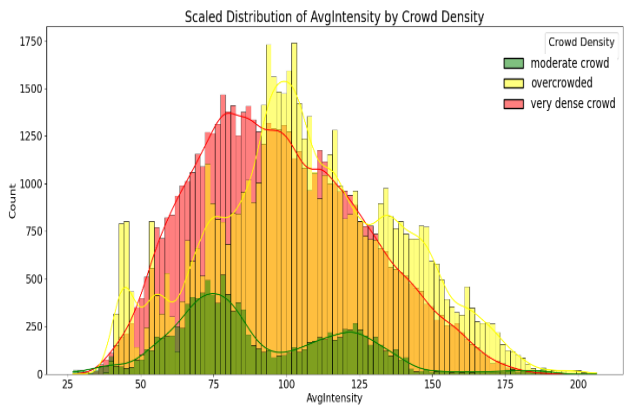

Fig. 6. Distribution of the "Average Intensity" (AvgIntensity) feature across crowd density categories.

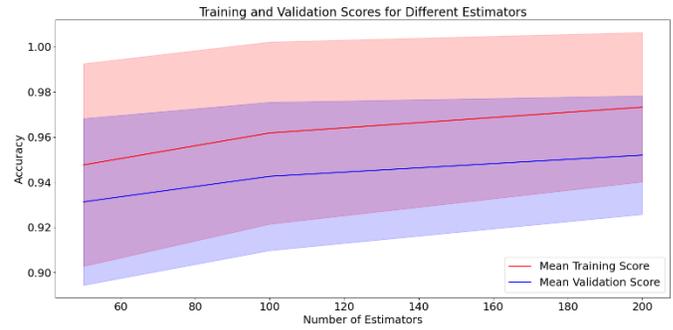

Fig. 7. Training and validation scores.

Finally, Fig. 7 shows the training and validation accuracy trends as the number of estimators increases, along with shaded regions representing the standard deviation for each. Both training and validation accuracy improve as the number of estimators increases and the model demonstrates stable performance without significant overfitting, as indicated by the relatively small gap between training and validation scores.

### A. Model Performance and Confusion Matrix Analysis

Metrics including accuracy, precision, recall and F1-score were used to evaluate the proposed model. Fig. 8 shows the results of these metrics as 87% accuracy, 93% precision, 87% recall, and 84% F1-score, which therefore indicate the model's high reliability, especially in identifying frames with very dense crowd conditions, which is crucial for real-time monitoring systems.

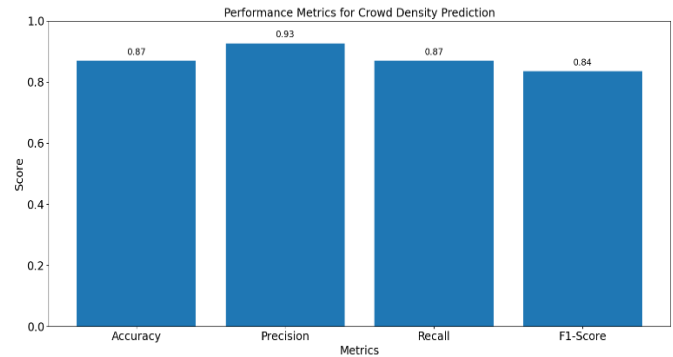

Fig. 8. Performance metrics.

In addition, a confusion matrix was constructed to analyze misclassifications. The matrix revealed that majority of misclassifications occurred between moderate crowd and overcrowded. While, frames classified as very dense crowd showed minimal false negatives thereby ensuring effective detection in critical situations. Fig. 9 presents the confusion matrix and provides a detailed breakdown of predictions versus true labels.

Moreover, an error percentage or misclassification rate based on the values of confusion matrix using the following equation:

$$Error\ Percentange = \frac{Total\ Misclassified\ Instances}{Total\ Instances} \times 100$$

$$Error\ Percentage = \frac{94}{4382} \times 100 \approx 2.14\%$$

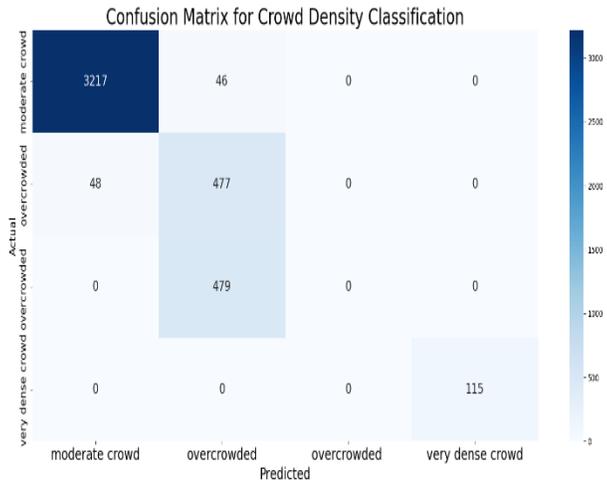

Fig. 9. Confusion matrix.

The 2.14% error percentage is ideal for high precision in a critical application such as classifying the surveillance videos of the holy sites during Hajj, especially in determining very dense crowd conditions, as false alerts or false estimations of the crowd situation could lead to either unnecessary triggering of emergency efforts or disasters. Furthermore, the model triggered visual alerts represented as red overlays for frames classified as 'very dense crowd'. This visual feedback could enhance the practical usability of the system in high-risk environments like Hajj.

The results demonstrate the success of the proposed approach in addressing crowd density classification challenges with high accuracy and low error percentage. The use of GridSearchCV for hyperparameter optimization and data augmentation for feature enhancement contributed significantly to the model's success. However, some limitations have been noted such as misclassification between moderate and overcrowded conditions, which could indicate the need for further enhancement to the methods used for extracting the features of the video frames. Also, real-time processing performance could benefit from hardware acceleration or further enhancement of the model.

Future work may look into contextual data integration such as temporal crowd flow patterns. In addition, extending the system to support multi-camera feeds could improve scalability for large-scale events like Hajj.

*B. Comparsion with Previous Studies*

The proposed model has been compared with other models in previous studies reviewed in Section II based on specific key evaluation criteria as shown in Table I. These criteria were collected from the previously reviewed studies. It was found that the proposed model in this study outperforms other models in its tailored feature extraction, real-time usability, focus on crowd density, and specific application to Hajj which make it uniquely suited for improving crowd management during Hajj. Although the proposed model achieves an accuracy of 87% which is slightly lower than the reported accuracy in some studies, that outperforms others in its targeted application of crowd density classification during Hajj. Unlike models that focus on detecting individual anomalies or specific behaviors, the proposed model is specifically designed to classify crowd density into actionable categories (moderate, overcrowded, very dense) and provide real-time visual alerts for very dense conditions. That is essential during Hajj to avoid crises such as the Mina stampede, which tragically led to the deaths of 1,000 pilgrims [21], and the collapse of a crawler crane at the Holy Mosque, resulting in 107 deaths and over 230 injuries [22] due to the dense crowds. That makes it more practical and reliable for real-world crowd management during Hajj.

Furthermore, with a low error rate of 2.14% and a balanced performance across metrics (precision: 93%, recall: 87%, F1-score: 84%), the proposed model ensures robust and consistent results tailored to Hajj-specific challenges.

TABLE I. COMPARISON OF PROPOSED MODEL WITH RELATED WORKS ON KEY EVALUATION CRITERIA (LEGEND √ MEANS INCLUDED/STRONGLY ADDRESSED, × MEANS NOT ADDRESSED AND ≈ MEANS PARTIALLY INCLUDED)

| Criteria | Tailored Feature Extraction | Real-Time Visual Alerts | Low Misclassification Rate | Focus on Crowd Density | Application to Hajj-Specific Scenarios |
|---|---|---|---|---|---|
| Bhuiyan [4] | × | × | × | ≈ | √ |
| Aldayri and Albattah [11] | × | × | × | × | × |
| Bhuiyan et al.[12] | × | × | × | × | ≈ |
| Nasir et al.[13] | × | × | × | × | ≈ |
| Alhothali et al.[14] | ≈ | × | × | × | ≈ |
| Habib et al.[15] | × | √ | × | × | × |
| Alafif et al. [16] | × | × | × | × | ≈ |
| Miao et al.[19] | √ | × | × | × | √ |
| Alafif et al.[20] | √ | × | × | × | √ |
| Proposed model | √ | √ | √ | √ | √ |

*C. Discussion*

Even though this study uses the HAJJv2 dataset, which represents different crowd scenarios from various holy sites, such as Tawaf, Sa'i, Mina and Arafat, thereby covering different crowd density levels, biases may exist due to the under representation of certain density levels or specific scenarios, such as moderately crowded conditions during less busy times. To address this, data augmentation techniques were applied to balance the dataset and reduce classification bias. The model's performance on unseen data indicates its generalizability, though further validation with external datasets could provide additional insights into fairness.

The observed misclassifications between moderate and overcrowded categories highlight areas for improvement. Introducing additional contextual features, such as temporal crowd patterns or environmental conditions may enhance fairness and reduce boundary-level classification errors.

The proposed model in this study demonstrates trustworthiness by providing interpretable outputs based on clear and relevant features like area, edge density and spatial coordinates. These features align with real-world crowd dynamics, as evidenced by the strong correlations observed in very dense crowd situations, and the real-time visual alerts represented as red overlays for very dense crowd frames, further enhancing trust by offering actionable and transparent feedback to decision-makers.

The proposed model's low error rate of 2.14% combined with high precision 93% and recall 87% metrics ensures its reliability in critical scenarios, especially in avoiding false negatives for very dense crowd conditions. That makes the system suitable for real-time applications during high-stakes events like Hajj.

Studies like Bhuiyan et al. [12] and Alhothali et al. [14] focused on detecting individual anomalies, which differ from the comprehensive crowd density classification targeted by this study which uniquely combines tailored feature extraction with advanced machine learning techniques. Also, the proposed model balances computational efficiency with an accuracy of 87% offering a practical solution for detecting hazardous crowds during Hajj. While hybrid approaches as shown by Miao et al. [19] and Alafif et al. [20] which integrate deep learning techniques (e.g., CNNs, ResNet-50, YOLOv2) with traditional methods (e.g., Random Forests, Kalman filters) have demonstrated high accuracy in abnormal behavior detection, their reliance on computationally intensive models, advanced infrastructure (e.g., UAVs, edge-cloud models) and heavily annotated datasets often limits their scalability and real-time applicability during Hajj.

Finally, to further improve fairness and trustworthiness, future work may incorporate temporal crowd flow data, environmental factors and multi-camera feeds to enhance model robustness and scalability. Validation with external datasets can help address potential biases and improve generalizability, ensuring broader applicability to diverse crowd management scenarios.

## V. CONCLUSION

This study proposed a model to classify crowd density in video surveillance recorded at the holy sites during Hajj into three classifications: moderate crowd, overcrowded and very dense crowd. The videos were segmented into frames, and features such as Edge Density, LBP texture, and area were extracted, which were identified as critical predictors for crowd density classification. Data augmentation and class balancing were employed to enhance the training dataset. A GBC was chosen for its robustness in handling imbalanced datasets and its ability to capture complex relationships between features. GridSearchCV was utilized to optimize the classification process, tuning key parameters such as the number of estimators, learning rate and maximum depth through cross-validation to achieve optimal performance.

The findings illustrate the capability of the proposed model to accurately classify crowd density levels. This proves its usefulness for real-time monitoring and managing crowds in critical locations.